\documentclass[10pt,twocolumn,letterpaper]{article}

\usepackage[pagenumbers]{cvpr} 

\usepackage{graphicx}
\usepackage{amsmath}
\usepackage{amssymb}
\usepackage{booktabs}

\usepackage[pagebackref,breaklinks,colorlinks]{hyperref}

\usepackage[capitalize]{cleveref}
\crefname{section}{Sec.}{Secs.}
\Crefname{section}{Section}{Sections}
\Crefname{table}{Table}{Tables}
\crefname{table}{Tab.}{Tabs.}

\def\cvprPaperID{*****} 
\def\confName{CVPR}
\def\confYear{2022}

\begin{document}

\title{Data Augmentation for Text-based Person Retrieval \\
	Using Large Language Models}

\author{Zheng Li, Lijia Si, Caili Guo, Yang Yang, Qiushi Cao \\
	Beijing University of Posts and Telecommunications \\
	{\tt\small \{lizhengzachary, silijia, guocaili, yangyang01, 819972578\}@bupt.edu.cn} \\
}
\maketitle

\begin{abstract}
	Text-based Person Retrieval~(TPR) aims to retrieve person images that match the description given a text query.
	The performance improvement of the TPR model relies on high-quality data for supervised training.
	However, it is difficult to construct a large-scale, high-quality TPR dataset due to expensive annotation and privacy protection.
	Recently, Large Language Models~(LLMs) have approached or even surpassed human performance on many NLP tasks, creating the possibility to expand high-quality TPR datasets.
	This paper proposes an LLM-based Data Augmentation~(LLM-DA) method for TPR.
	LLM-DA uses LLMs to rewrite the text in the current TPR dataset, achieving high-quality expansion of the dataset concisely and efficiently.
	These rewritten texts are able to increase the diversity of vocabulary and sentence structure while retaining the original key concepts and semantic information.
	In order to alleviate the hallucinations of LLMs, LLM-DA introduces a Text Faithfulness Filter~(TFF) to filter out unfaithful rewritten text.
	To balance the contributions of original text and augmented text, a Balanced Sampling Strategy~(BSS) is proposed to control the proportion of original text and augmented text used for training.
	LLM-DA is a plug-and-play method that can be easily integrated into various TPR models.
	Comprehensive experiments on three TPR benchmarks show that LLM-DA can improve the retrieval performance of current TPR models.
\end{abstract}

\section{Introduction}
\label{sec:intro}
\label{sec:formatting}
Text-based Person Retrieval~(TPR)~\cite{jiang2023cross} aims to retrieve person images that match the description given a text query, which is a sub-task of image-text retrieval~\cite{chen2020imram} and person re-identification~(Re-ID)~\cite{ye2021deep}.
TPR can assist in identifying individuals captured in surveillance footage based on textual descriptions.
TPR has implications for surveillance and security applications, where identifying individuals based on textual descriptions can aid in law enforcement and public safety efforts.

Current studies~\cite{jiang2023cross, bai2023rasa} on TPR mainly focus on extracting discriminative feature representations and fine-grained feature alignment to achieve competitive retrieval performance.
As a multi-modal learning task, the performance improvement of the TPR model relies on high-quality data for supervised training.
However, it is difficult to construct a large-scale, high-quality TPR dataset for TPR model training.
There are two reasons: 
1) \textbf{Lack of data.} 
Due to privacy protection, it is difficult to obtain large-scale person images. 
2) \textbf{Lack of high-quality annotation.}
Text annotation is tedious and inevitably introduces annotator biases.
Therefore, the texts in the current TPR datasets are usually short and cannot comprehensively describe the characteristics of the target person.
In order to solve this problem, Yang~\textit{et al.}~\cite{yang2023towards} construct a large-scale multi-attribute dataset, MALS, for the pre-training of the TPR task.
It takes a lot of manpower and material resources to construct MALS, and we are grateful for their contribution to the TPR field.

\begin{figure}[t]
	\centering
	\includegraphics[width=\linewidth]{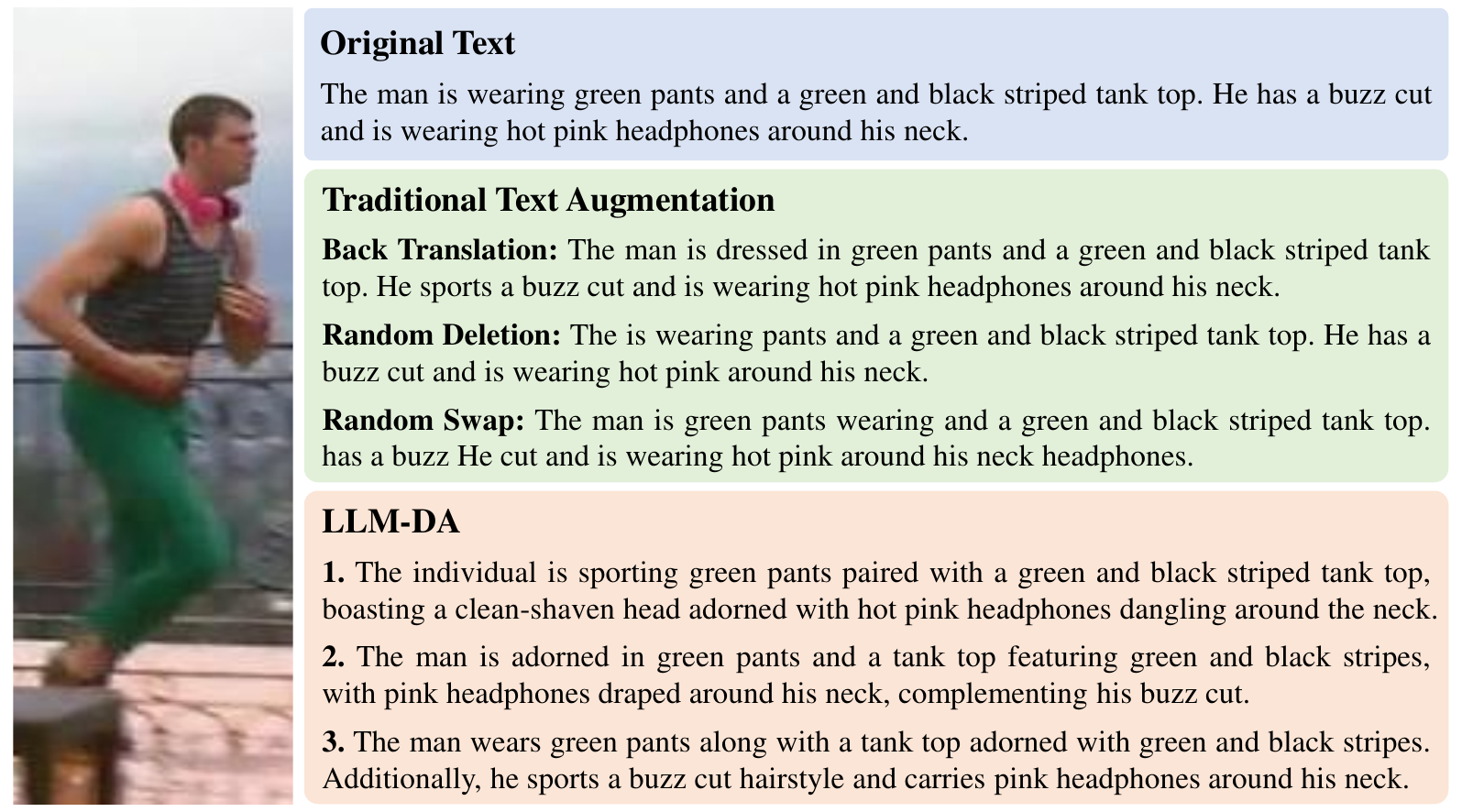}
	\caption{Original person image, original text, and augmented text.}
	\label{example}
\end{figure}
In addition to constructing large-scale datasets, data augmentation is also an effective way to expand data scale and facilitate model training.
Compared with dataset construction, data augmentation has lower labor and material costs.
Cao~\textit{et al.}~\cite{cao2024empirical} conduct a comprehensive empirical study on data augmentation in the TPR task, including image augmentation and text augmentation.
Image augmentation methods include traditional removal and alteration.
Text augmentation methods include back translation, random deletion, \textit{etc}.
Most of these traditional image augmentation methods can improve the retrieval performance of TPR models.
However, we find that these traditional text augmentation methods do not significantly improve retrieval performance, and some methods even reduce retrieval performance.
These simple text augmentation methods have limited improvement in text diversity.
More seriously, some crude text augmentation methods, such as random deletion and random swap, can destroy the correct sentence structure and even change the original semantic concept of the text, as shown in \figurename~\ref{example}.
These low-quality augmented texts can have a negative impact on model training.

Recently, Large Language Models~(LLMs) have approached or even surpassed human performance on many NLP tasks, creating the possibility to expand high-quality TPR datasets.
LLM can be used to rewrite the original text to generate new text, thereby achieving text augmentation.
Thanks to the powerful semantic understanding and generation capabilities of LLMs, these rewritten texts are able to increase the diversity of vocabulary and sentence structure while retaining the original key concepts and semantic information.
\figurename~\ref{example} shows the augmented text we generated using the open-source LLM Vicuna~\cite{vicuna2023}. 
The augmented text generated by LLM can enhance the diversity of the text while maintaining the correct sentence structure.
Although LLM has powerful generation capabilities, hallucinations have always been a thorny problem that LLM cannot solve. It is possible for LLM to generate augmentation text that does not meet expectations, which is an issue that needs to be addressed.

In this paper, we propose an LLM-based Data Augmentation~(LLM-DA) method for TPR.
LLM-DA uses LLMs to rewrite the text in the current TPR dataset, achieving high-quality expansion of the dataset concisely and efficiently.
These rewritten texts are able to increase the diversity of vocabulary and sentence structure while retaining the original key concepts and semantic information.
In order to alleviate the hallucinations of LLMs, LLM-DA introduces a Text Faithfulness Filter~(TFF) to filter out unfaithful rewritten text.
To balance the contributions of original text and augmented text, a Balanced Sampling Strategy~(BSS) is proposed to control the proportion of original text and augmented text used for training.
LLM-DA neither changes the original model architecture nor affects the form of the original loss function. 
Therefore, LLM-DA is a plug-and-play method that can be easily integrated into various TPR models.
The major contributions of this paper are summarized as follows:
\begin{itemize}
	\item
	We propose an LLM-based Data Augmentation~(LLM-DA) method for TPR, using LLM to rewrite the text in the current TPR dataset, achieving high-quality expansion of the dataset concisely and efficiently.
	This is the first exploration of using LLM for data augmentation in the TPR task.
	\item
	We propose a Text Faithfulness Filter~(TFF) to filter out unfaithful rewritten text to alleviate the hallucinations of LLMs.
	\item
	We propose a Balanced Sampling Strategy~(BSS) to control the proportion of original text and augmented text used for training.
	\item
	LLM-DA can be plug-and-play integrated into various TPR models.
	Comprehensive experiments on three TPR benchmarks show that LLM-DA can improve the retrieval performance of current TPR models.
\end{itemize}
\section{Related work}
\subsection{Text-based Person Retrieval}
Text-based Person Retrieval~(TPR)~\cite{jiang2023cross} aims to retrieve person images that match the description given a text query.
Li~\textit{et al.}~\cite{li2017person} first propose TPR, which is a sub-task of image-text retrieval~\cite{chen2020imram} and person re-identification~(Re-ID)~\cite{ye2021deep}.
Feature extraction and feature alignment are the core steps to TPR.
Current studies on TPR mainly focus on these two aspects.

\textbf{Feature Extraction} refers to extracting discriminative features from input person images and text descriptions.
Li~\textit{et al.}~\cite{li2017identity, li2017person} use Long Short-Term Memory~(LSTM) to extract text features and Convolution Neural Networks~(CNN) to extract image features.
Zhu~\textit{et al.}~\cite{zhu2021dssl} use the ResNet-50~\cite{he2016deep} pretrained on the ImagNet dataset to extract image features and the Bidirectional Gate Recurrent Unit~(Bi-GRU) to extract text features.
In recent years, with the emergence of Transformer ~\cite{vaswani2017attention} and Bidirectional Encoder Representations from Transformers~(BERT) ~\cite{devlin2018bert}, large-scale pre-trained models have been gradually used to extract features.
Han~\textit{et al.}~\cite{han2021text} first introduce Contrastive Language-Image Pre-Training (CLIP)~\cite{radford2021learning} for feature extraction in TPR.
Jiang~\textit{et al.}~\cite{jiang2023cross} use CLIP image and text encoders to extract image and text features, respectively.
Yang~\textit{et al.}~\cite{yang2023towards} apply Swin Transformer~\cite{liu2021swin} to extract image features and BERT to extract text features. 
Bai~\textit{et al.}~\cite{bai2023rasa} use the large-scale vision-language pre-trained model ALBEF~\cite{li2021align} to extract image and text features.

\textbf{Feature Alignment} refers to the process of effectively matching image and text features.
Li~\textit{et al.}~\cite{li2017identity} use Cross-Modal Cross-Entropy~(CMCE) loss for feature alignment. 
Li~\textit{et al.}~\cite{li2017person} propose a Recurrent Neural Network with Gated Neural Attention~(GNA-RNN) mechanism to capture the relationship between images and text.
In addition to loss functions and attention mechanisms, recent studies~\cite{zhu2021dssl, niu2020improving, wang2020vitaa, jing2020cross} use more complex models for feature alignment.
Zhu \textit{et al.}~\cite{zhu2021dssl} use five different modules and loss functions for feature alignment to make full use of multi-modal and multi-granular information to improve retrieval performance.
Niu \textit{et al.}~\cite{niu2020improving} propose a Multi-granularity Image-text Alignment~(MIA) model to alleviate the cross-modal fine-grained problem.
The Visual-Textual Attribute Alignment Model (ViTAA) module is proposed to align person partial features and textual attribute features with a k-reciprocal sampling alignment loss~\cite{wang2020vitaa}.
Jing \textit{et al.}~\cite{jing2020cross} propose a Moment Alignment Network~(MAN) to solve the cross-domain and cross-modal alignment problems.
Later studies focus more on the fine-grained alignment of multimodalities.
By designing an implicit relation reasoning module in the random mask language paradigm, Jiang \textit{et al.}~\cite{jiang2023cross} complete the fine-grained alignment of modalities and achieve cross-modal text and visual interaction.
Yang \textit{et al.}~\cite{yang2023towards} incorporate the tasks of Image-Text Contrastive Learning~(ITC), Image-Text Matching Learning~(ITM), and Masked Language Modeling~(MLM) to impose the alignment constraints.
Bai \textit{et al.}~\cite{bai2023rasa} propose Relationship-Aware~(RA) learning and Sensitivity-Aware~(SA) learning.
RA focuses on the correlation between images and text,which is a coarse-grained optimization.
SA focuses more on the interaction between images and text, which is fine-grained optimization.

Looking back at the development of TPR, most studies focus on improving retrieval performance through the feature level, but high-quality data is crucial to improving the performance of supervised learning models.
Privacy protection and annotation make building large-scale, high-quality datasets challenging.
In order to solve this problem, Yang~\textit{et al.}~\cite{yang2023towards} construct a large-scale multi-attribute dataset, MALS, for the pre-training of the TPR task, which takes a lot of manpower and material resources. 
In order to obtain large-scale, high-quality data at a low cost, this paper considers introducing data augmentation into TPR.

\subsection{Data Augmentation}
Data augmentation increases the diversity of the data and improves the robustness of the model by changing and expanding the original data.
TPR datasets are usually constructed in the form of image-text pairs.
Therefore, the data augmentation of TPR datasets requires considering both image augmentation and text augmentation.

\textbf{Image Augmentation.}
There are a lot of methods of image augmentation.
Commonly used traditional methods include random cropping, flipping, scaling, color transformation, \textit{etc}.
In addition, some novel image augmentation methods, such as Mixup~\cite{zhang2017mixup} and CutMix~\cite{yun2019cutmix}, are also widely used.
Mixup randomly selects two images in each batch and mixes them in a certain ratio to generate a new image.
CutMix generates a new image by randomly cutting and pasting image fragments from different areas of the image, thereby improving the ability of models to learn local features.
Previous studies~\cite{krizhevsky2017imagenet, simonyan2014very, he2016deep, szegedy2016rethinking, sajjadi2016regularization} have demonstrated that the data augmentation of images can effectively improve the generalization and robustness of the model.
Cao \textit{et al.}\cite{cao2024empirical} point out that image augmentation has a certain effect on improving the performance of TPR.

\textbf{Text Augmentation.}
Compared with image augmentation, text augmentation faces more challenges because of the complexity, abstraction, flexibility, scarcity, and diversity of text.
Easy Data Augmentation~(EDA)~\cite{wei2019eda} is a simple text augmentation method, including synonym replacement, random insertion, random swap, and random deletion.
Back translation~\cite{fadaee2017data} generates new sentences by translating text into another language and then back.
Although back translation is widely used and has achieved certain success, due to cultural differences between different languages, it may lead to semantic inconsistency and is not universal.
CutMixOut~\cite{fawakherji2024textaug} combines Cutout~\cite{devries2017improved} and CutMix~\cite{yun2019cutmix} to randomly replace and remove text subsequences through a binary mask.
However, these methods may destroy the structural and semantic information of sentences, and the augmented texts lack diversity.
With the widespread application of LLMs, text augmentation can be performed using LLMs.
While ensuring the semantic integrity of the sentence, LLMs can also increase the diversity of sentence structure and form, effectively enhancing the generalization and robustness of the model.
Such as Fan \textit{et al.}~\cite{fan2024improving} improve CLIP performance by augmenting text with LLMs.

\begin{figure*}[t]
	\centering
	\includegraphics[width=\linewidth]{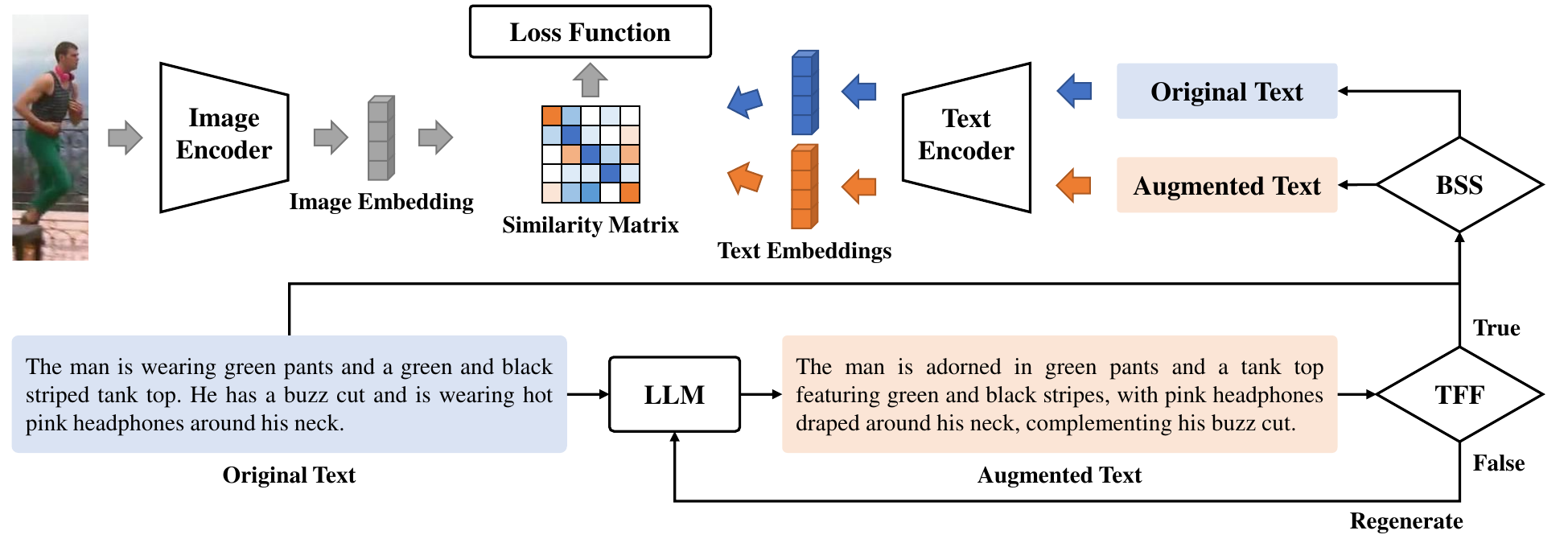}
	\caption{The framework for using LLM-based Data Augmentation~(LLM-DA) in TPR model training.}
	\label{train}
\end{figure*}
\subsection{Large Language Models}
The Transformer architecture provides the basis for the subsequent generation of LLMs.
Radford~\textit{et al.}~\cite{radford2018improving} introduce the Generative Pretrained Transformer~(GPT) model, which is based on the Transformer architecture and serves as the foundation for the advancement of LLMs.
Subsequently, the emergence of a series of GPT models~\cite{radford2019language, brown2020language, achiam2023gpt} further promotes the development of this field.
Moreover, the release of open-sourced models like LLaMA~\cite{touvron2023llama} and GLM~\cite{du2022glm}, fine-tuned for various tasks, has served as the backbone for numerous applications.
Vicuna~\cite{vicuna2023} introduces a more economical option with its 7B and 13B versions while maintaining impressive performance, contributing significantly to the progress in the field of LLMs.
These models collectively achieve comparable performances across various benchmarks, creating the possibility to expand high-quality TPR datasets.

Although LLMs can perform well on many different tasks, there are still some problems that need to be solved when applying LLMs for text augmentation. 
One of the key issues is the hallucination of LLMs. 
The hallucination refers to the situation where the grammatical correctness, fluency, and authenticity of the generated text are inconsistent with the original input text or even inconsistent with the facts~\cite{ye2023cognitive}. 
The hallucination problem not only reduces the reliability of generated text but may also lead to an uneven quality of output text and sometimes even abnormal text. 
Therefore, it is necessary to slove the hallucination of LLMs.


\section{Methodology}
\subsection{Preliminary}
Text-based person retrieval~(TPR) is defined as retrieving person images relevant to the description of a given text query.
We denote $ \mathcal{V} = \{ V_{i} \}_{i=1}^{I} $ as a collection of person images and $ \mathcal{T} = \{ T_{i} \}_{i=1}^{I} $ as a collection of text descriptions, where $ V_{i} $ is a person image and $ T_{i} $ is a text description.
In TPR, given a text description $ T_{i} $, the goal is to find the most relevant person image $ V_{i} $ from the person image collection $ \mathcal{V} $.
Current TPR models generally follow a common framework, which contains an image encoder $ \boldsymbol{f}_{img}(\cdot) $ and a text encoder $ \boldsymbol{f}_{text}(\cdot) $.
The similarity $ s(V_{i}, T_{i}) $ between $ V_{i} $ and $ T_{i} $ is computed based on the encoded image feature $ \boldsymbol{f}_{img}(V_{i}) $ and text feature $ \boldsymbol{f}_{text}(T_{i}) $.
Finally, the retrieval results are obtained by ranking the similarities.

\subsection{LLM-based Data Augmentation}
\figurename~\ref{train} shows the framework for using LLM-based Data Augmentation~(LLM-DA) in TPR model training.
LLM-DA first utilizes an LLM to rewrite the original text to generate augmented text.
Then, in order to alleviate the hallucinations of LLMs, LLM-DA introduces a Text Faithfulness Filter~(TFF) to filter out unfaithful rewritten text.
On the one hand, the faithfully rewritten text is used as augmented text for model training.
On the other hand, LLM-DA discards the unfaithful rewritten text and uses LLM again to rewrite the original text to generate augmented text.
Finally, in order to balance the contributions of original text and augmented text, LLM-DA introduces a Balanced Sampling Strategy~(BSS) to control the proportion of original text and augmented text used for training through sampling.
Through the BSS, the caculated similarity matrix between person images and texts is a mixed similarity matrix, which contains both the similarity between the image and the original text and the similarity between the image and the augmented text.
This mixed similarity matrix is used to calculate the loss function and implement model training.

\begin{figure}[t]
	\centering
	\includegraphics[width=\linewidth]{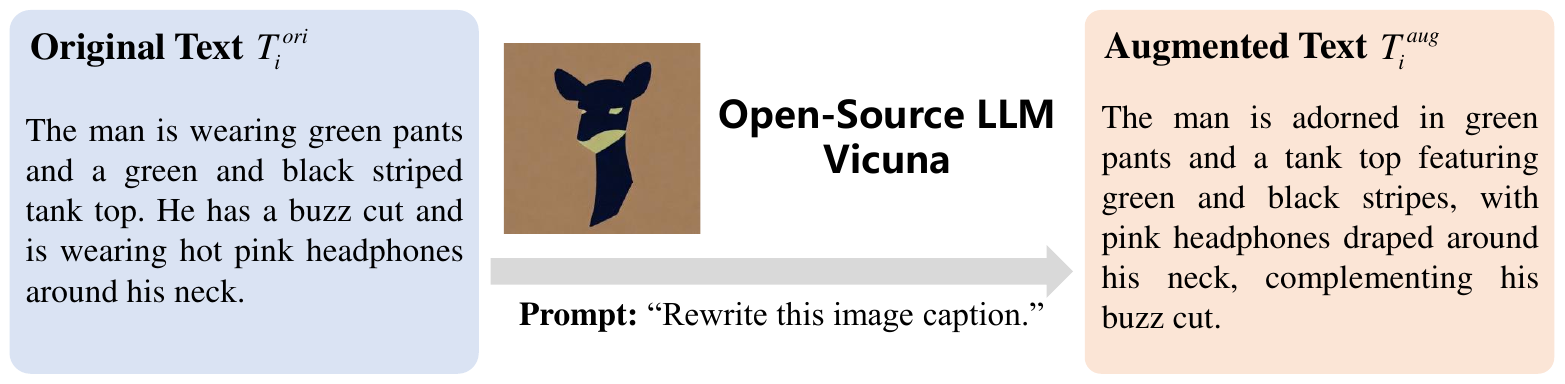}
	\caption{Using LLM for text augmentation.}
	\label{llm}
\end{figure}
\figurename~\ref{llm} shows how to use LLMs to generate augmented text.
In this paper, we choose the LLM Vicuna~\cite{vicuna2023} for text augmentation, which is an open-source chatbot trained by fine-tuning LLaMA on user-shared conversations collected from ShareGPT. 
Preliminary evaluation using GPT-4 as a judge shows Vicuna achieves more than 90\% of the quality of OpenAI ChatGPT and Google Bard.
We concatenate the original text $ T_{i}^{ori} $ and prompt ``\textit{Rewrite this image caption.}'' and enter them into Vicuna together.
Vicuna rewrites the original text $ T_{i}^{ori} $ and returns the augmented text:
\begin{equation} \label{rewrite}
	T_{i}^{aug} 
	= \text{LLM}(
	\text{Concat}
	(T_{i}^{ori}, \text{Prompt})
	).
\end{equation}
Thanks to the powerful generalization of LLMs, most of the text rewritten using LLMs can maintain the same key concepts and semantic information as the original text.
In addition, with the powerful generation capabilities of LLMs, using LLMs to rewrite text can enrich the diversity of text data.

\subsection{Text Faithfulness Filter}
\begin{figure}[t]
	\centering
	\includegraphics[width=0.9\linewidth]{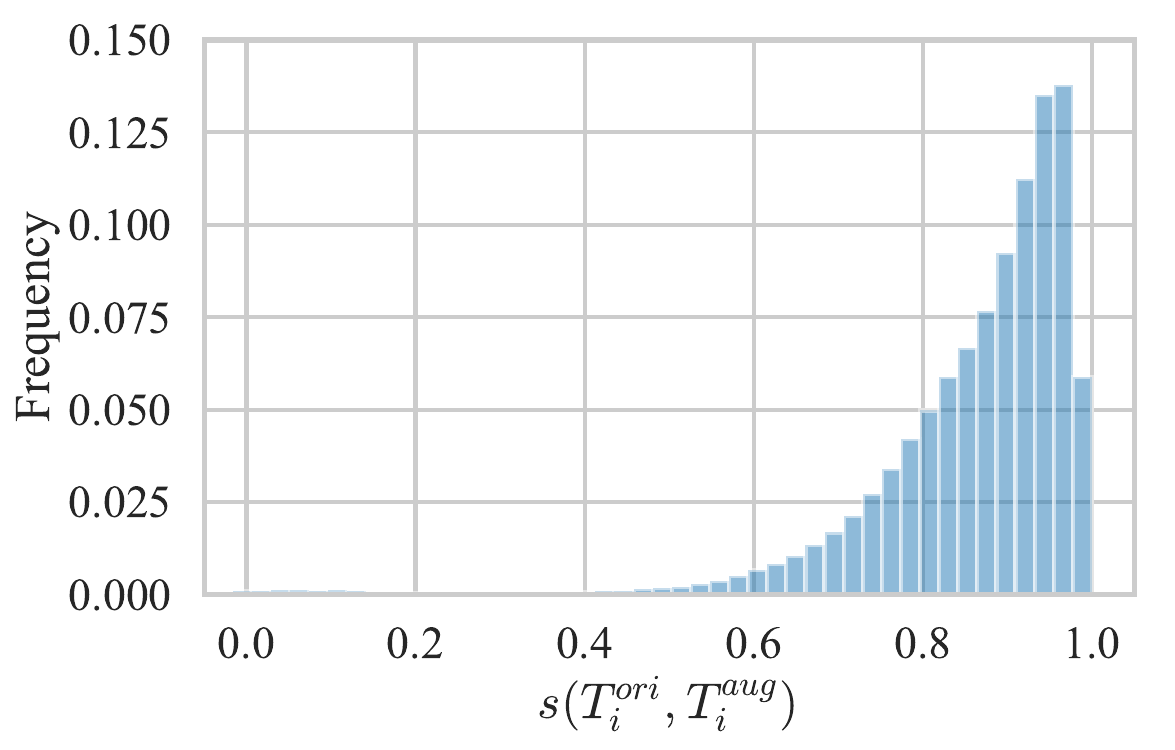}
	\caption {Distribution of $ s(T_{i}^{ori}, T_{i}^{aug}) $ on the CUHK-PEDES dataset.}
	\label{distrubution}
\end{figure}
Although LLMs have demonstrated powerful capabilities in various tasks, hallucination is still a problem that LLMs have not completely solved.
In the process of using LLMs for text augmentation, we find that the rewritten text output by LLMs may not be semantically consistent with the original text, and LLMs may even output text in other languages or garbled characters.
We calculated the semantic similarity between the original text and the augmented text, as shown in \figurename~\ref{distrubution}. 
More than 90\% of the augmented text has a semantic similarity greater than 0.6 with the original text. 
But there are still a small number of augmented texts that are semantically inconsistent with the original texts.
In order to alleviate the hallucinations of LLMs, LLM-DA introduces a Text Faithfulness Filter~(TFF) to filter out unfaithful rewritten text.

\begin{figure}[t]
	\centering
	\includegraphics[width=\linewidth]{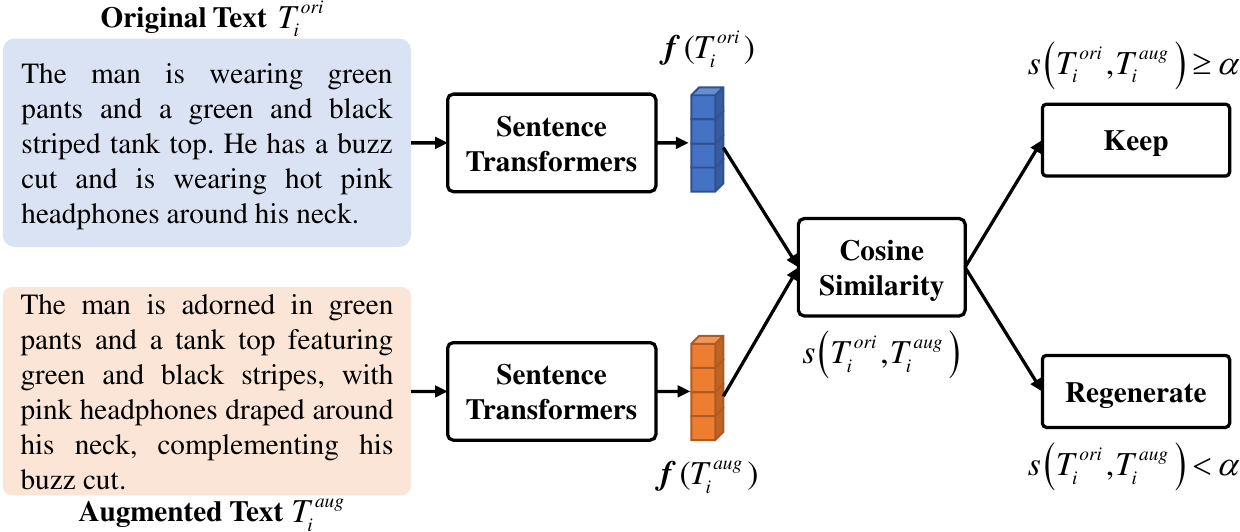}
	\caption{Text Faithfulness Filter~(TFF).}
	\label{tff}
\end{figure}
The architecture of TFF is shown in \figurename~\ref{tff}.
The purpose of TTF is to filter out augmented text that does not match the semantics of the original text.
Therefore, there is a need to measure the semantic similarity between the original text and the augmented text.
To this end, we introduce the Sentence Transformers framework to implement semantic similarity calculation.
Sentence Transformers is a Python framework for state-of-the-art sentence, text and image embeddings. 
First, we use Sentence Transformers $ \boldsymbol{f}_{st}(\cdot) $ to encode the original text $ T_{i}^{ori} $ and augmented text $ T_{i}^{aug} $ to obtain original text embedding $ \boldsymbol{f}_{st}(T_{i}^{ori}) $ and augmented text embedding $ \boldsymbol{f}_{st}(T_{i}^{aug}) $.
Then, the semantic similarity between the original text and the augmented text can be calculated using simple cosine similarity:
\begin{equation}
	s(T_{i}^{ori}, T_{i}^{aug})
	= 
	\dfrac
	{
		\boldsymbol{f}_{st}(T_{i}^{ori})^{\top} 
		\cdot
		\boldsymbol{f}_{st}(T_{i}^{aug})
	}
	{
		\left \Vert
		\boldsymbol{f}_{st}(T_{i}^{ori}) 
		\right \Vert
		\left \Vert
		\boldsymbol{f}_{st}(T_{i}^{aug})
		\right \Vert
	}.
\end{equation}
We set a similarity threshold $ \alpha $. 
When $ s(T_{i}^{ori}, T_{i}^{aug}) < \alpha $, the augmented text is considered to be semantically inconsistent with the original text.
LLM-DA discards the unfaithful rewritten text and uses LLM again to rewrite the original text to generate augmented text.
When $ s(T_{i}^{ori}, T_{i}^{aug}) \geq \alpha $, the augmented text is considered to be semantically consistent with the original text.
The faithfully rewritten text is used as an augmented text for model training.
Through TFF filtering, noise data in augmented text can be effectively removed, and the quality of training data can be improved.


\subsection{Balanced Sampling Strategy}
After obtaining the augmented text, the simplest way to use the augmented text for training is to directly add the augmented text to the original dataset.
However, there may still be a small amount of noise data in the augmented text, which can have a negative impact on model training.
In addition, the distribution of augmented text may be different from that of original text.
Introducing too much augmented text for training may be detrimental to the generalization of the model.
Therefore, in order to balance the contributions of original text and augmented text, LLM-DA introduces a Balanced Sampling Strategy~(BSS) to control the proportion of original text and augmented text used for training through sampling.

We define $ T_{i}^{*} $ as the text ultimately used for training. 
The process of BSS can be expressed as:
\begin{equation}
	T_{i}^{*} =
	\begin{cases}
		T_{i}^{ori} , & r_{i} > \beta, \\
		T_{i}^{aug} , & r_{i} \leq \beta,
	\end{cases}
\end{equation}
where $ r_{i} $ is a random number following a uniform distribution with a value range of $ [0, 1] $.
$ \beta $ is a predefined sampling threshold hyperparameter used to control the proportion of original text and augmented text for training.
Balancing the contributions of original text and augmented text can reduce the interference of noisy data on model training while increasing the diversity of training data.

Through the BSS, the caculated similarity matrix between person images and texts is a mixed similarity matrix:
\begin{equation}
	\mathbf{S}
	= \begin{bmatrix} 
		s(V_{1}, T_{1}^{*})  &  \dots & s(V_{N}, T_{1}^{*}) \\
		\vdots  &  \ddots  & \vdots \\ 
		s(V_{1}, T_{N}^{*})  &  \dots & s(V_{N}, T_{N}^{*})
	\end{bmatrix},
\end{equation}
where $ N $ is the batch size. 
$ \mathbf{S} $ contains both the similarity $ s(V_{i}, T_{i}^{ori}) $ between the image and the original text and the similarity $ s(V_{i}, T_{i}^{aug}) $ between the image and the augmented text.
This mixed similarity matrix is used to calculate the loss function and implement model training.
In this paper, we use CLIP as a baseline model to implement TPR. 
The contrastive learning loss used by CLIP after applying LLM-DA can be written as:
\begin{equation}
	\mathcal{L}_{\text{Contrastive}}
	^{v \rightarrow t}
	= - \sum_{i=1}^{N}
	\log 
	\dfrac
	{
		\exp
		(s(V_{i}, T_{i}^{*}) / \tau)
	}
	{
		\sum_{j=1}^{N} 
		\exp(s(V_{i}, T_{j}^{*}) / \tau) 
	}, 
\end{equation}
where $ \tau $ is a temperature coefficient.
$ \mathcal{L}_{\text{Contrastive}}^{v \rightarrow t} $ is the loss of image-to-text retrieval, and the loss $ \mathcal{L}_{\text{Contrastive}}^{t \rightarrow v} $ of text-to-image retrieval is symmetrical to $ \mathcal{L}_{\text{Contrastive}}^{v \rightarrow t} $.
LLM-DA neither changes the original model architecture nor affects the form of the original loss function. 
Therefore, LLM-DA is a plug-and-play method that can be easily integrated into various TPR models.

\section{Experiments}
\subsection{Experimental Setup}
\textbf{Datasets.}
We conduct comprehensive experiments on three TPR datasets: CUHK-PEDES~\cite{li2017person}, ICFG-PEDES~\cite{ding2021semantically}, and RSTPReid~\cite{zhu2021dssl}.
\begin{itemize}
	\item 
	\textbf{CUHK-PEDES}~\cite{li2017person} is the first dataset dedicated to TPR, which contains 40,206 images and 80,412 textual descriptions for 13,003 identities.
	Following the official data split, the training set consists of 11,003 identities, 34,054 images, and 68,108 textual descriptions.
	The validation set and test set contain 3,078 and 3,074 images, 6158 and 6156 textual descriptions, respectively, and both of them have 1,000 identities.
	\item 
	\textbf{ICFG-PEDES}~\cite{ding2021semantically} contains a total of 54,522 images for 4,102 identities. 
	Each image has only one corresponding textual description. 
	The dataset is divided into a training set and a test set; the former comprises 34,674 image-text pairs of 3,102 identities, while the latter contains 19,848 image-text pairs for the remaining 1,000 identities.
	\item 
	\textbf{RSTPReid}~\cite{zhu2021dssl} contains 20,505 images of 4,101 identities from 15 cameras.
	Each identity has five corresponding images taken by different cameras, and each image is annotated with two textual descriptions. 
	Following the official data split, the training, validation, and test sets contain 3,701, 200, and 200 identities, respectively.
\end{itemize}

\textbf{Evaluation Metrics.}
We adopt the popular Rank-K metrics~(K = 1, 5, and 10) as the primary evaluation metrics.
Rank-K reports the probability of finding at least one matching person image within the top-K candidate list when given a textual description as a query. 
In addition, for a comprehensive evaluation, we also adopt the mean Average Precision~(mAP) as another retrieval criterion.
The higher Rank-K and mAP indicate better performance.

\textbf{Implementation Details.}
Our all experiments are conducted on an NVIDIA GeForce RTX 3090 GPU using PyTorch.
We use CLIP as a baseline model to implement TPR.
CLIP is a neural network trained on a variety of image-text pairs.
Many TPR methods use CLIP as the backbone of the model.
Since this paper mainly focuses on data augmentation, in order to reflect the gains of data augmentation, we do not use the various tricks proposed for TPR and only use the original CLIP for experiments.
CLIP-ViT-B/16 and CLIP-ViT-B/32 are used as the image encoders, and CLIP Text Transformer is used as the text encoder.
All person images are resized to $ 224 \times 224 $. 
The maximum length of the textual token sequence is set to $ 77 $.
The model is trained with the AdamW optimizer with a learning rate initialized to $ 1 \times 10^{-5} $.
The training batch size is $ 80 $.
We use an early stopping strategy to select the optimal model.
When the mAP of five consecutive epochs after an epoch no longer grows, the model saved in this epoch is selected as the final model for subsequent testing.

\begin{table}[t]
	\small
	\setlength\tabcolsep{6pt}
	\begin{center}
		\begin{tabular}{lcccccccccccc}
			\toprule[1pt]
			Method & Rank-1 & Rank-5 & Rank-10 & mAP \\
			\hline
			
			\specialrule{0em}{2pt}{0pt}
			CLIP~(ViT-B/32) &  60.82 & 81.47 & 88.50 & 54.51 \\
			\textbf{+ LLM-DA} & \textbf{61.45} & \textbf{82.41} & \textbf{88.68} & \textbf{54.77} \\
			\hline
			
			\specialrule{0em}{2pt}{0pt}
			CLIP~(ViT-B/16) &  64.59 & 83.59 & 89.51 & 58.02\\
			\textbf{+ LLM-DA} & \textbf{66.47} & \textbf{85.32} & \textbf{91.03} & \textbf{59.93} \\
			
			\bottomrule[1pt]
		\end{tabular}
	\end{center}
	\caption{Experimental results on the CUHK-PEDES dataset.}
	\label{CUHK-PEDES}
\end{table}
\subsection{Improvements to TPR Models}
In this section, we present the performance improvements of three TPR datasets on two baseline models.
We use two CLIP models used in the latest TPR research~\cite{cao2024empirical} as baseline models.

\textbf{Improvements on the CUHK-PEDES Dataset.}
\tablename~\ref{CUHK-PEDES} shows the experimental results on the CUHK-PEDES dataset.
The performance after applying LLM-DA is better than the original baseline on both models.
The performance improvement on the more powerful CLIP~(ViT-B/16) model is more significant than that of the CLIP~(ViT-B/32) model.
Specifically, after applying LLM-DA, the retrieval performance metrics Rank-1 and mAP can be improved by 2.91\% and 3.29\%, respectively, compared with the original CLIP~(ViT-B/32).

\begin{table}[t]
	\small
	\setlength\tabcolsep{6pt}
	\begin{center}
		\begin{tabular}{lcccccccccccc}
			\toprule[1pt]
			Method & Rank-1 & Rank-5 & Rank-10 & mAP \\
			\hline
			
			\specialrule{0em}{2pt}{0pt}
			CLIP~(ViT-B/32) &  51.40 & 77.05 & 84.95 & 41.21\\
			\textbf{+ LLM-DA} & \textbf{52.15} & \textbf{77.65} & \textbf{85.00} & \textbf{41.57} \\
			\hline
			
			\specialrule{0em}{2pt}{0pt}
			CLIP~(ViT-B/16) &  55.75 & 80.20 & 88.20 & 44.73\\
			\textbf{+ LLM-DA} & \textbf{58.70} & \textbf{81.20} & \textbf{88.35} & \textbf{45.93} \\
			
			\bottomrule[1pt]
		\end{tabular}
	\end{center}
	\caption{Experimental results on the RSTPReid dataset.}
	\label{RSTPReid}
\end{table}
\textbf{Improvements on the RSTPReid Dataset.}
\tablename~\ref{RSTPReid} shows the experimental results on the RSTPReid dataset.
On both models, the performance after applying LLM-DA is superior to the initial baseline.
Similar to the performance on the CUHK-PEDES dataset, the performance improvement on the more powerful CLIP~(ViT-B/16) model is more significant than the CLIP~(ViT-B/32) model.
In particular, compared to the original CLIP~(ViT-B/32), the retrieval performance metrics Rank-1 and mAP are improved by 3.50\% and 2.68\%, respectively, after applying LLM-DA.

\begin{table}[t]
	\small
	\setlength\tabcolsep{6pt}
	\begin{center}
		\begin{tabular}{lcccccccccccc}
			\toprule[1pt]
			Method & Rank-1 & Rank-5 & Rank-10 & mAP \\
			\hline
			
			\specialrule{0em}{2pt}{0pt}
			CLIP~(ViT-B/32) &  52.75 & 72.27 & 79.52 & 31.29 \\
			\textbf{+ LLM-DA} & \textbf{53.04} & \textbf{72.58} & \textbf{79.84} & \textbf{32.00} \\
			\hline
			
			\specialrule{0em}{2pt}{0pt}
			CLIP~(ViT-B/16) &  56.70 & 75.25 & 81.55 & 35.20\\
			\textbf{+ LLM-DA} & \textbf{58.05} & \textbf{75.43} & \textbf{81.74} & \textbf{37.33} \\
			
			\bottomrule[1pt]
		\end{tabular}
	\end{center}
	\caption{Experimental results on the ICFG-PEDES dataset.}
	\label{ICFG-PEDES}
\end{table}
\textbf{Improvements on the ICFG-PEDES Dataset.}
\tablename~\ref{ICFG-PEDES} shows the experimental results on the CUHK-PEDES dataset.
Applying LLM-DA improves performance on both models over the baseline.
In particular, Rank-1 and mAP retrieval performance metrics are improved by 2.38\% and 6.05\%, respectively, following the application of LLM-DA in comparison to the initial CLIP~(ViT-B/32).
In summary, LLM-DA can improve the performance of all metrics on all three datasets.
This demonstrates the generalization of LLM-DA.

\subsection{Comparisons with Text Data Augmentation Methods}
LLM-DA is a text augmentation method. 
There are many traditional text augmentation methods used in TPR:
\begin{itemize}
	\item 
	\textbf{Random Deletion} randomly removes words from text.
	\item 
	\textbf{Random Swap} randomly selects two words from the text and swaps their positions.
	\item 
	\textbf{Back Translation} translates the original text into a specific language and back again.
\end{itemize}
We compare LLM-DA with the above traditional text augmented methods.
For back translation, we use French as the intermediate language.
It has a relatively closer form to English and introduces fewer changes to the translated back text in semantics than other languages.

\tablename~\ref{da} shows the performance comparisons with traditional text augmentation methods on the RSTPReid dataset.
LLM-DA shows significant performance gains compared with other text augmentation methods.
LLM-DA significantly outperforms the baseline on all evaluation metrics.
However, several other traditional text augmentation methods may fall below the baseline on some evaluation metrics.
Random deletion may remove keywords from the text. 
Random swap may change the original grammatical structure of the text.
Both methods may destroy the correct sentence structure and even change the original semantic concept of the text, which may have a negative impact on model training.
Back translation can maintain the semantic concepts and grammatical structure of the original text, but the text diversity it can increase is relatively limited.
LLM-DA utilizes the powerful generalization and generation capabilities of LLMs, which can not only maintain the semantic concepts and grammatical structure of the original text but also significantly improve the text diversity, thus achieving the most significant performance gain.
\begin{table}[t]
	\small
	\setlength\tabcolsep{6pt}
	\begin{center}
		\begin{tabular}{lcccccccccccc}
			\toprule[1pt]
			Method  & Rank-1 & Rank-5 & Rank-10 & mAP  \\
			\hline
			
			\specialrule{0em}{2pt}{0pt}
			CLIP~(ViT-B/16) & 55.75 & 80.20 & 88.20 & 44.73 \\
			+ Random Deletion & 56.50 & 80.05 & 88.00 & 44.13  \\
			+ Random Swap & 56.95 & 80.05 & 88.25 & 45.13 \\
			+ Back Translation &  55.95 & 80.85 & \textbf{88.50} & 45.17 \\   
			\textbf{+ LLM-DA} & \textbf{58.85} & \textbf{81.10} & 88.35 & \textbf{46.13} \\
			
			\bottomrule[1pt]
		\end{tabular}
	\end{center}
	\caption{Comparisons with traditional text augmentation methods on the RSTPReid dataset.}
	\label{da}
\end{table}

\subsection{Ablation Study}
\begin{table}[t]
	\small
	\setlength\tabcolsep{6pt}
	\begin{center}
		\begin{tabular}{ccccccccccccc}
			\toprule[1pt]
			DA & TFF & BSS & Rank-1 & Rank-5 & Rank-10 & mAP \\
			\hline
			
			\specialrule{0em}{2pt}{0pt}
			- & - & - & 64.59 & 83.59 & 89.51 & 58.02 \\
			\checkmark & - & - &  64.78 & 84.06 & 89.93 & 58.95\\
			\checkmark & \checkmark & - & 65.66 & 85.14 & 90.98 & 59.17 \\
			\checkmark & - & \checkmark & 64.94 &  84.29 & 90.59 & 58.12 \\
			\checkmark & \checkmark & \checkmark & \textbf{66.33} & \textbf{85.41} & \textbf{91.03} & \textbf{59.92} \\
			
			\bottomrule[1pt]
		\end{tabular}
	\end{center}
	\caption{Ablation studies on the CUHK-PEDES dataset.}
	\label{ablation}
\end{table}
\textbf{Impact of Different Modules.}
LLM-DA mainly consists of three components: LLM-based Data Augmentation~(DA), Text Faithfulness Filter~(TFF) and Balanced Sampling Strategy~(BSS).
DA first utilizes an LLM to rewrite the original text to generate augmented text.
Then, in order to alleviate the hallucinations of LLMs, TFF filters out unfaithful rewritten text.
Finally, in order to balance the contributions of original text and augmented text, BSS controls the proportion of original text and augmented text used for training through sampling.

\tablename~\ref{ablation} shows the impact of different modules in LLM-DA.
The experiment is conducted on the CUHK-PEDES dataset.
We adopt the CLIP~(ViT-B/16) model as the baseline for the experiment.
Compared with the baseline, only data augmentation of text can improve retrieval performance, but the performance improvement is not significant.
After TFF filtering, the retrieval performance is significantly improved. 
This is because TFF filters out augmented text that is inconsistent with the semantic concepts of the original text, reduces the noise in the training data, and alleviates the negative impact of noisy data on model training.
There is a little improvement in retrieval performance following BSS sampling.
This is because balancing the proportion of original text and augmented text can also alleviate the negative impact of noisy data to a certain extent and improve the generalization of the model.
Combining the three modules can achieve optimal retrieval performance. This shows that the three modules introduced by LLM-DA can not only improve performance individually but also complement each other.

\begin{figure}
	\centering
	\begin{subfigure}{0.495\linewidth}
		\includegraphics[width=\linewidth]{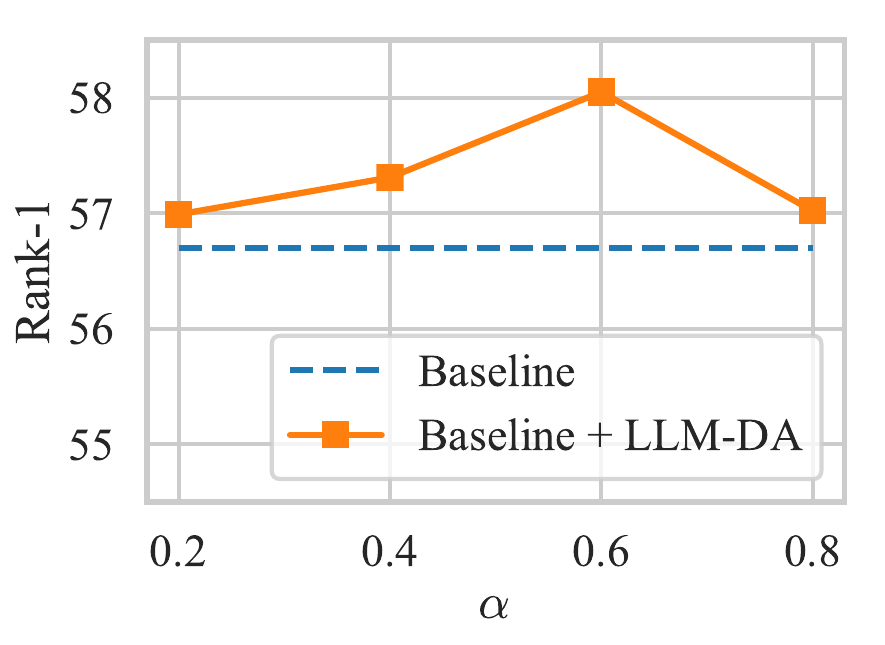}
		\caption{Rank-1}
		\label{alpha_rank1}
	\end{subfigure}
	\hfill
	\begin{subfigure}{0.495\linewidth}
		\includegraphics[width=\linewidth]{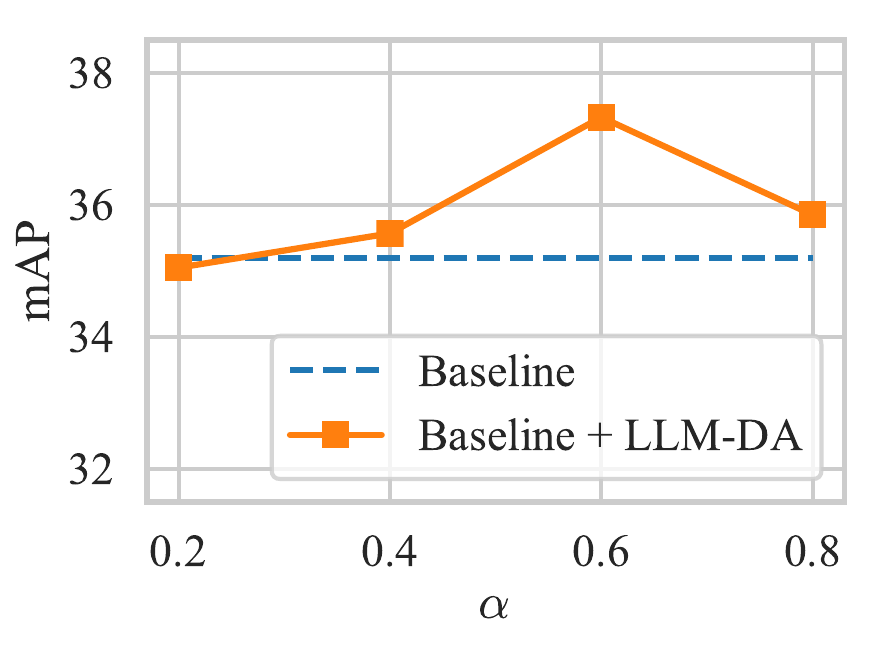}
		\caption{mAP}
		\label{alpha_map}
	\end{subfigure}
	\caption{The impact of hyperparameter $ \alpha $ on retrieval performance on the ICFG-PEDES dataset.}
	\label{alpha}
\end{figure}
\begin{figure}
	\centering
	\begin{subfigure}{0.495\linewidth}
		\includegraphics[width=\linewidth]{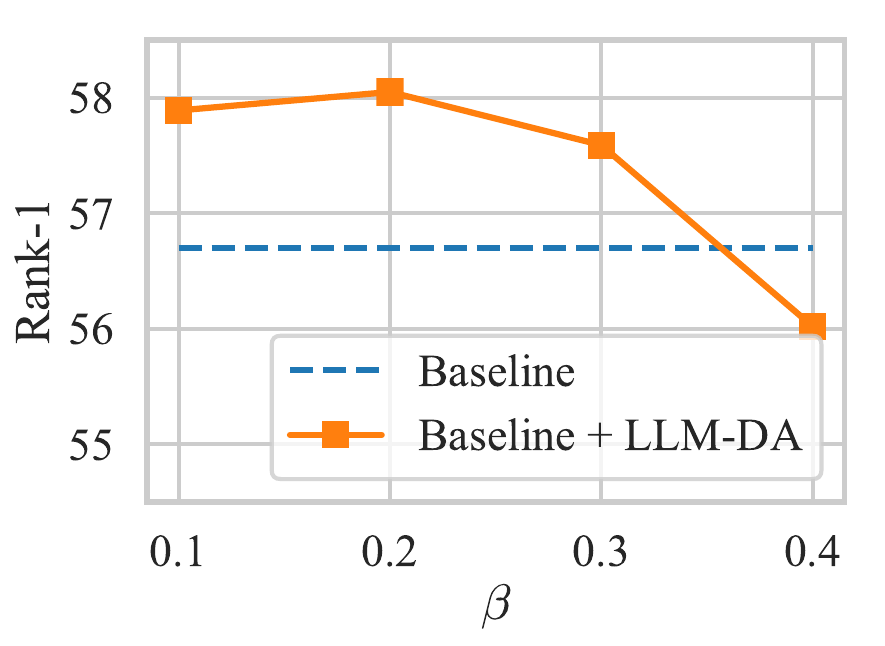}
		\caption{Rank-1}
		\label{beta_rank1}
	\end{subfigure}
	\hfill
	\begin{subfigure}{0.495\linewidth}
		\includegraphics[width=\linewidth]{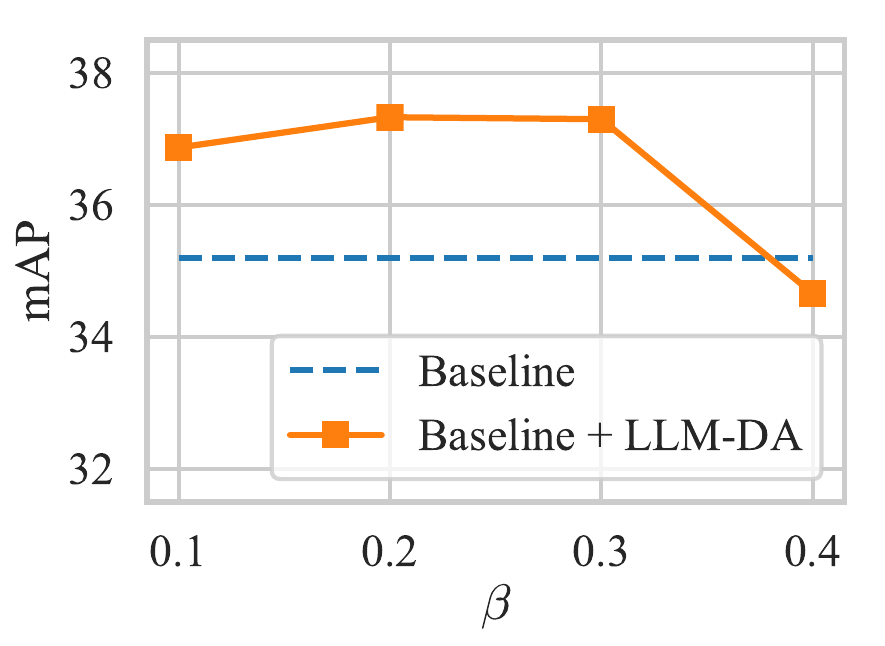}
		\caption{mAP}
		\label{beta_map}
	\end{subfigure}
	\caption{The impact of hyperparameter $ \beta $ on retrieval performance on the ICFG-PEDES dataset.}
	\label{beta}
\end{figure}
\begin{figure*}[t]
	\centering
	\includegraphics[width=\linewidth]{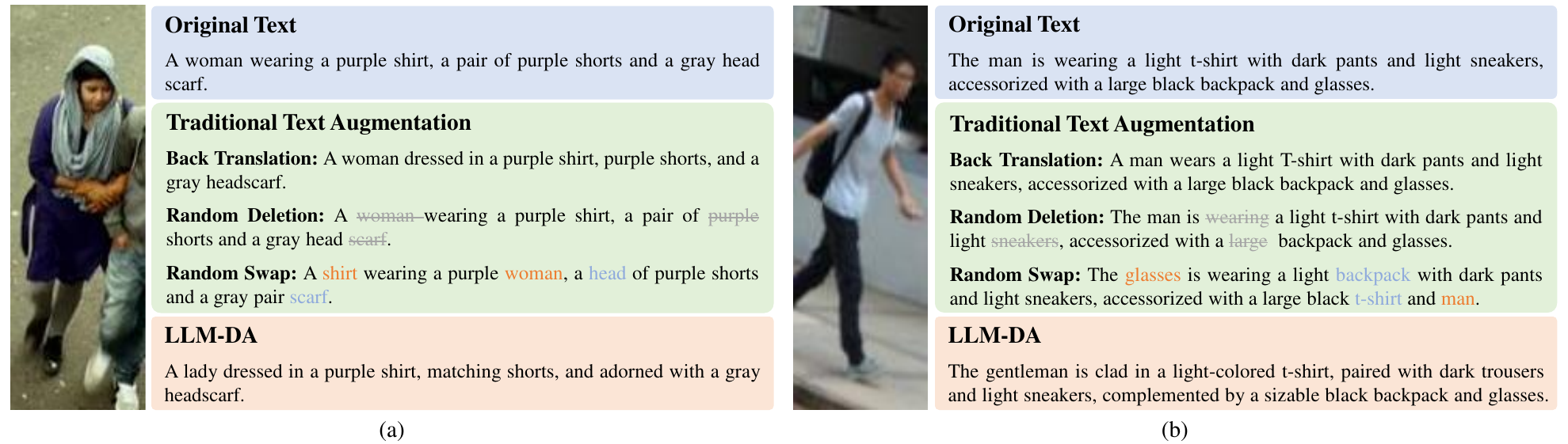}
	\caption{Qualitative results of different text data augmentation methods on the CUHK-PEDES dataset.}
	\label{qualitative}
\end{figure*}
\textbf{Hyperparameter Analysis.}
There are two hyperparameters~($ \alpha $ and $ \beta $) in LLM-DA that can be tuned. 
$ \alpha $ is the predefined similarity threshold hyperparameter in TFF, which is used to decide whether the augmented text should be retained for training.
$ \beta $ is a predefined sampling threshold hyperparameter in BSS, which is used to control the proportion of original text and augmented text for training.
We experiment with several hyperparameter settings on the ICFG-PEDES dataset using the CLIP~(ViT-B/16) model.

As shown in \figurename~\ref{alpha}, as $ \alpha $ increases, the retrieval performance first increases and then decreases. 
At $ \alpha < 0.4 $, LLM-DA does not significantly improve retrieval performance since more noisy data is used for training, which has a negative impact on model performance.
When $ \alpha = 0.6 $, the retrieval performance reaches the optimal level. 
However, a bigger $ \alpha $ is not always better. 
When $ \alpha > 0.8 $, since the augmented text is similar to the original text, the diversity of the text data is insufficient and the retrieval performance is reduced, which is not conducive to the generalization of the model.
Therefore, the choice of $ \alpha $ requires a trade-off between reducing noise data and increasing the diversity of text data.

As shown in \figurename~\ref{beta}, as $ \beta $ increases, the retrieval performance first increases and then decreases.
When the value of $ \beta $ is small, only less augmented text participates in training, and the contribution to model performance improvement is not significant.
When $ \beta = 0.2 $, the retrieval performance reaches the optimal level. 
When $ \beta > 0.3 $, the retrieval performance drops significantly.
There are two reasons why the performance decreases when the value of $ \beta $ is large.
On the one hand, there may still be a small amount of noise data in the augmented text, which has a negative impact on model training.
On the other hand, the distribution of augmented text may be different from the distribution of the original text.
To sum up, the value of $ \beta $ needs to balance the proportion of original text and augmented text participating in training.

\subsection{Qualitative Results}
\figurename~\ref{qualitative} presents the qualitative results of different text data augmentation methods on the CUHK-PEDES dataset.
We compare the proposed LLM-DA method with three traditional text augmention methods.
Text augmented using traditional methods may destroy the semantic concepts of the original text.
In addition, these texts are similar to the sentence structure of the original text and lack diversity.
On the other hand, the text augmented by LLM-DA has more complete semantics and richer sentence structure than the traditional method.
This shows that the LLM-DA method has significant advantages in text augmentation, can better retain the semantic information of the original text, and can generate more natural and fluent sentences.

\section{Conclusion}
In this paper, we propose an LLM-based Data Augmentation~(LLM-DA) method for Text-based Person Retrieval~(TPR).
Specifically, we use LLMs to rewrite the text in the current TPR dataset, achieving high-quality expansion of the dataset concisely and efficiently.
In order to alleviate the hallucinations of LLMs, we introduce a Text Faithfulness Filter~(TFF) to filter out unfaithful rewritten text.
To balance the contributions of original text and augmented text, a Balanced Sampling Strategy~(BSS) is proposed to control the proportion of original text and augmented text used for training.
LLM-DA is a plug-and-play method that can be easily integrated into various TPR models and improve their retrieval performance.
In future work, we plan to expand LLM-DA to more cross-modal retrieval tasks.



{\small
	\bibliographystyle{ieee_fullname}
	\bibliography{egbib}
}

\end{document}